\documentclass[letterpaper, 10 pt, conference]{ieeeconf}  
\IEEEoverridecommandlockouts 
\usepackage{graphicx} \usepackage{times}
\usepackage{amsmath}
\usepackage{amssymb}
\usepackage{bm}
\usepackage{setspace}
\usepackage{mathtools} 
\usepackage{algorithm}
\usepackage{booktabs}
\usepackage[noend]{algpseudocode}
\usepackage{epsfig} \usepackage{epstopdf}
\usepackage{multirow} \usepackage{subfigure}
\usepackage{caption} \usepackage{cite}
\usepackage{color}	\usepackage{url} 
\usepackage{gensymb} % \usepackage{mnsymbol}
\usepackage{balance}
\usepackage[dvipsnames]{xcolor}
	\newtheorem{example}{Example}

\renewcommand{\baselinestretch}{0.98}

\title{ \bf
Semi-Autonomous Teleoperation via \\
Learning Non-Prehensile Manipulation Skills
}
\author{Sangbeom Park$^{\dag}$, 
Yoonbyung Chai$^{\dag}$, 
Sunghyun Park$^{\dag}$,
Jeongeun Park$^{\dag}$,
Kyungjae Lee$^{\ddag}$,
and Sungjoon Choi$^{\dag}$
\thanks{This work was supported by Samsung Electronics (IO201230-08278-01) and Institute of Information \& Communications Technology Planning \& Evaluation (IITP) grant funded by the Korea government (MSIT) (No. 2019-0-00079, Artificial Intelligence Graduate School Program (Korea University)).}
\thanks
{$^{\dag}$Sangbeom Park, Yoonbyung Chai, Sunghyun Park, Jeongeun Park, and Sungjoon Choi are with 
the Department of Artificial Intelligence, 
Korea University, Seoul, Korea (e-mail: {\tt\scriptsize
\{psb5657, 
yoonbyung-chai, 
sunghyun-park, 
baro0906, 
sungjoon-choi\}@korea.ac.kr}).}%
\thanks{$^{\ddag} $Kyungjae Lee is with 
the Department of Artificial Intelligence, 
Chung-ang University, Seoul, Korea (e-mail: {\tt\scriptsize kyungjae.lee@ai.cau.ac.kr}).}%
}
\begin{document}
\maketitle % \thispagestyle{empty} \pagestyle{empty}

%
% Abstract
%
\begin{abstract}
In this paper, we present a semi-autonomous teleoperation framework for a pick-and-place task using an RGB-D sensor. In particular, we assume that the target object is located in a cluttered environment where both prehensile grasping and non-prehensile manipulation are combined for efficient teleoperation. A trajectory-based reinforcement learning is utilized for learning the non-prehensile manipulation to rearrange the objects for enabling direct grasping. From the depth image of the cluttered environment and the location of the goal object, the learned policy can provide multiple options of non-prehensile manipulation to the human operator. We carefully design a reward function for the rearranging task where the policy is trained in a simulational environment. Then, the trained policy is transferred to a real-world and evaluated in a number of real-world experiments with the varying number of objects where we show that the proposed method outperforms manual keyboard control in terms of the time duration for the grasping.
\end{abstract}

%
% Introdction
%
\section{Introduction}
% Importance of teleoperation
Autonomous robots, such as robot cleaners, have been gradually permeating into our everyday lives. However, when it comes to performing more complex tasks (e.g., picking up a specific object in a cluttered environment), fully autonomous robots may not be a preferable option due to their inability in perception and reasoning. In this regard, teleoperation may be a desirable choice. However, it is not straightforward to teleoperate a high degree-of-freedom robot using a limited control interface such as a keyboard.

% Shared autonomy + what we proposed 
To resolve this issue, a number of researches focus on shared autonomy for effective teleoperation. Shared autonomy addresses the problem of mediating different levels of autonomy and aims to find the appropriate level for the task. In \cite{Schilling_19}, the levels of autonomy are divided into \textit{intentions}, \textit{plans}, and \textit{selection of means}. In the context of pick-and-place tasks using a manipulator in a cluttered environment, \textit{intentions} may correspond to picking up a specific object, \textit{plans} may correspond to removing the blocking objects first and then picking up the target object, and \textit{selection of means} may correspond to controlling each joint of the robot, manually. In this work, we focus on providing the second level of shared autonomy (i.e., \textit{plans}) by letting operators to select high-level trajectories.

% What we propose
In this paper, we focus on picking up a specific target object using a robotic manipulator in a cluttered environment where directly picking up the target object may not be feasible due to the presence of blocking objects. In particular, we design the whole teleoperation system to iterate between two different modes, a prehensile grasping mode and a non-prehensile rearranging mode, where the users get to select the appropriate mode based on the sensory observation. For the grasping mode, we implement a simple yet effective interface for the users to simply select the target object within clustered objects from the  depth image. For the non-prehensile mode, we leverage trajectory-based reinforcement learning to train a policy that can provide multiple rearranging motions based on the current sensory observation and the location of the target object. The candidate end-effector trajectories are overlayed on the user screen. This allows the teleoperators to select their preferred option which plays a crucial role in our semi-autonomous teleoperation. 

% Our main contribution
Our main contributions of this work are twofold: We first present a semi-autonomous teleoperation framework that can provide high-level options through the appropriate level of shared autonomy. The other is to demonstrate our framework, which focused on \textit{plans} level using simulated and real-world experiments.
Following the user study, we show that our framework mitigate user difficulty to control manipulator and more effective in non-prehensile task. 
% Structure
\begin{figure*}[t]
 \centering
 \includegraphics[width=0.99\textwidth]{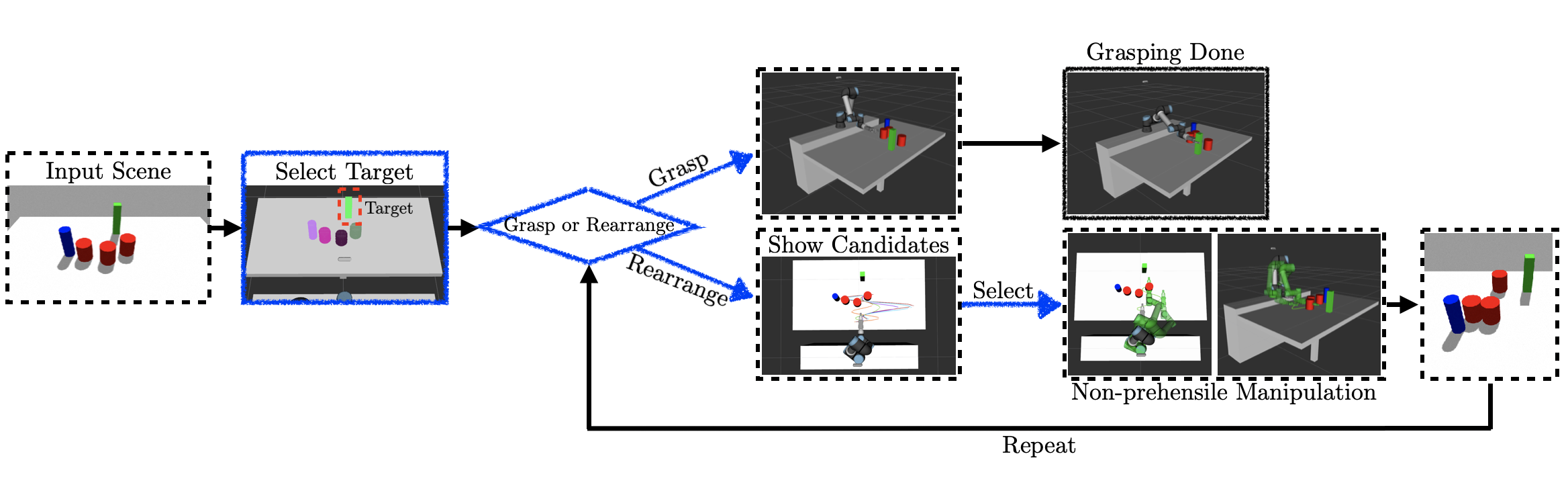}
 \vspace{-0.1pt}
 \caption{An overview of the proposed teleoperation framework where the blue colored boxes and arrow indicate the user's command. Given the current scene, the operator first selects the target object to grasp. Then, the operator decides whether to directly grasp the target or rearrange other objects. When the rearranging mode is selected, the user chooses the non-prehensile manipulation trajectories from the learned policy. 
 }
 \label{fig:overview}
 \vspace{-0.1pt}
\end{figure*}

%
% Related Work 
%
\section{Related Work} \label{sec:related_work}

%
% Shared Autonomy with Latent Models
%
\subsection{Shared Autonomy with Latent Models}

The recent work on teleoperation has proposed how to make a simple user interface using a latent space.
In \cite{Losey_20}, Losey et al. have investigated reducing the dimensionality of control inputs for efficient control of high dimensional robots.
While the method proposed in \cite{Losey_20} effectively learns a low dimensional latent action space and provides a low dimensional control interface, e.g., $1$-DoF latent action, however, the method excessively restricts the freedom of user where the participants do not clearly prefer the proposed method over the conventional teleoperation method due to the limited freedom in control.
In \cite{Jeon_20}, Jeon et al. have focused on learning and reflecting the human preferences in the low-dimensional latent action spaces.
In \cite{Karamcheti_21}, this method has been further extended into learning a context-based latent model that can provide a $1$-DoF control interface conditioned on a visual context.
While previous researches have advantages in learning a low-dimensional latent space, but they have the disadvantage of restricting the user's freedom.
In particular, since most existing studies have focused on mapping the robot's configuration to a latent space, it is difficult to visualize the entire sequence of the robot's behavior.
Such unseen and unpredictable interfaces make it uncomfortable for the user to teleoperate with high-dimensional robots.
For this reason, our framework proposes that the operator can decide a preferable trajectory considering the goal of the target during the task.

%
% Trajectory-based RL
%
\subsection{Trajectory-based Reinforcement Learning}

While reinforcement learning (RL) methods deal with a sequential decision-making problem, the policy of existing RL approaches \cite{Schulman_15, Schulman_17, Haarnoja_18} is often modeled to output an instantaneous action per each state observation. This approach will be referred to as instance-based reinforcement learning \cite{Choi_19_DLPG}. On the other hand, trajectory-based reinforcement learning directly models the high-dimensional trajectories of an agent given current context information. Deep latent policy gradient \cite{Choi_19_DLPG} utilizes the conditional variational autoencoder structure to accurately model the high-dimensional trajectory space where it has greatly shown its strength in terms of the sample efficiency with respect to the locomotion tasks. The main objective of utilizing reinforcement learning is not only to learn how to successfully rearrange objects without falling down but also has to provide appropriate and comprehensible joint trajectory candidates of the manipulator to the operating user. Hence, we utilize a trajectory-based RL algorithm.

%
% Non-prehensile Manipulation
%
\subsection{Non-prehensile Manipulation}
Most previous studies of rearranging objects to picking up a target object in cluttered environments have considered the non-prehensile planning method. 
In \cite{Lee_19}, Lee et al. have proposed the planning algorithm for rearranging the objects to grasp a target.
Furthermore, in \cite{Lee_21}, Lee et al. have focused on relocating the minimum number of objects based on tree search.
While several pieces of research have been conducted to reduce the planning time for rearranging blocking objects, the full non-prehensile manipulation might take excessive time for planning \cite{papallas_20, king_17, yuan_18}. 
However, it is essential to reduce a computational delay for real-time teleoperation. Hence, we employ learning-based sweeping motion to rearrange blocking obstacles instead of finding non-prehensile plans.

%
% Problem Formation 
%
\section{Problem Formulation} \label{sec:problem_formulation}

% Teleoperation of picking tasks 
The main objective of our work is to present an effective teleoperation method for picking a target object in a cluttered environment. In particular, we assume that the target object is blocked by obstacles where directly picking up the target may not be feasible. Traditionally, the blocking objects are removed by another pick-and-place task. However, we aim to use non-prehensile manipulation to rearrange the obstacles to speed up the picking up objective. The overall framework of the proposed method is illustrated in Figure \ref{fig:overview}. 

% Shared autonomy % TODO: add references, controlling assists robot--> user study part
In terms of shared autonomy, there exist multiple ways to formulate this problem. Perhaps, the most straightforward way of doing this would be manually controlling all the revolute joints of a manipulator, which corresponds to the undermost level of the shared autonomy (i.e., \textit{selection of means}). However, \cite{Freire_00} tells us that it would take an excessive amount of time to accomplish the picking task. 
% because of the complex interface, it would take an excessive amount of time to accomplish the picking task. 

% Shared Autonomy
On the other extreme, one might fully rely on the autonomous system where the planning module undertakes the picking tasks (i.e., task and motion planning (TAMP)) \cite{Nam_20}. While this approach could alleviate the burden of the teleoperator, possibly due to the cognitive limitation \cite{Liu_13}, an accurate physics-based simulation and perception modules are often necessary to successfully utilize TAMP in a real-world environment. Furthermore, the user preferences may not be reflected in the planned results. This may correspond to the uppermost level of shared autonomy (i.e., \textit{intention}). 

% What we present 
In this paper, we utilize the intermediate level of shared autonomy, \textit{plans}, where the users first get to choose whether to directly grasp the target object or to first rearrange the blocking objects by non-prehensile manipulation (i.e., push objects). Furthermore, when it comes to rearranging objects, our proposed method provides multiple but useful options to the operator so that the user selects the preferable one. 

% Learning how to rearrange objects
To this end, we utilize a trajectory-based reinforcement learning method to generate joint trajectories of a manipulator that can successfully and stably rearrange the obstacles while not having any contact with the target object. We also present a simple yet effective picking method based on the clustering of the depth data.

%
% Figure: Network
%
\begin{figure}[t]
 \centering
 \includegraphics[width=0.45\textwidth]{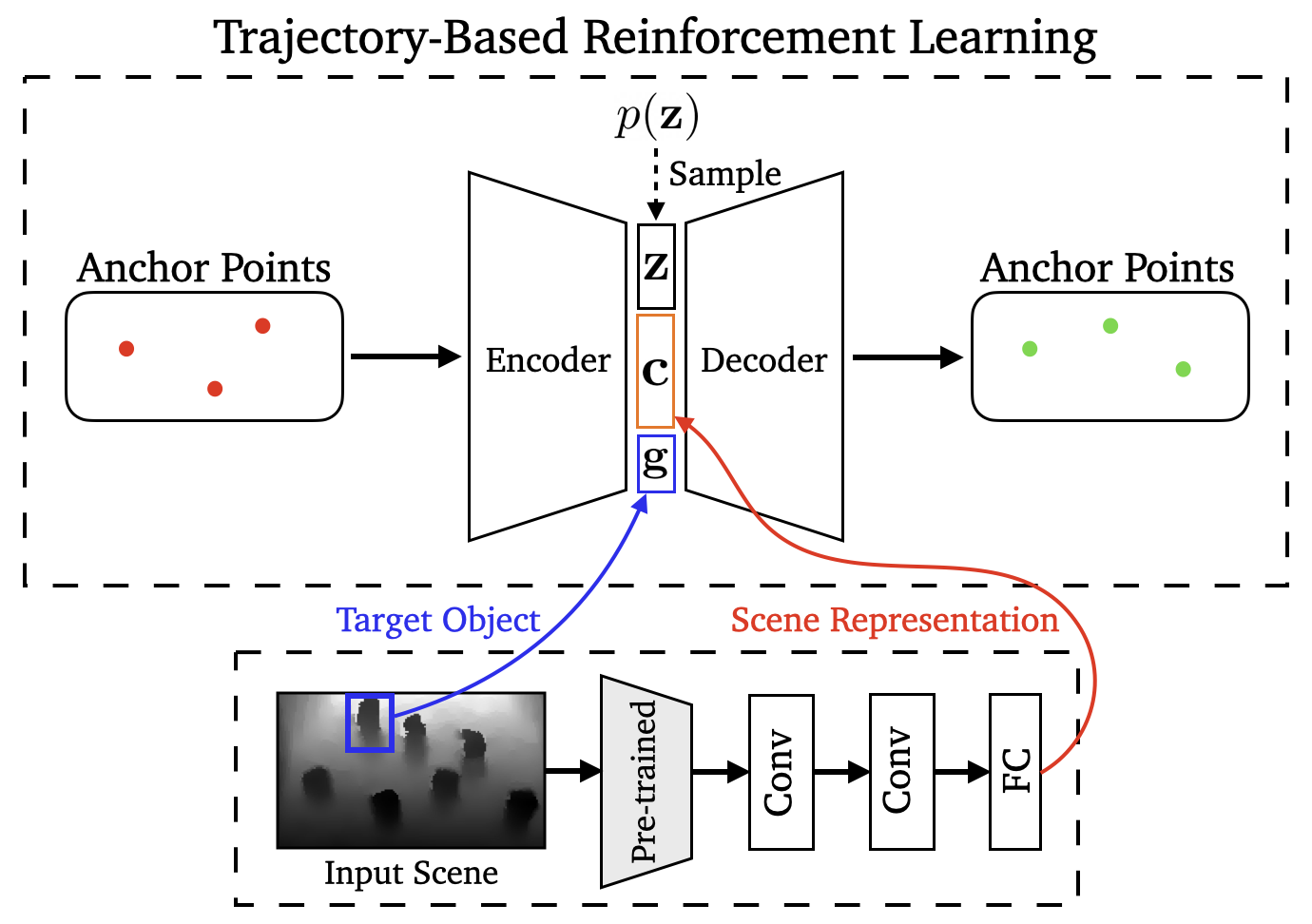}
 \vspace{-0.1pt}
 \caption{
 The network architecture of the proposed method. It consists of two networks: One is for the policy function of DLPG where the decoder part is used after the training phase and the other is for the representation of the depth image from the pre-trained CNN transferred from CAE. 
}
\label{fig:network}
\vspace{-0.1pt}
\end{figure}

%
% Proposed Method
%
\section{Proposed Method} \label{sec:proposed_method}

% Why we need a learning-based module for pushing task
Our proposed teleoperation system consists of two modules, a learning-based non-prehensile rearranging module in Section \ref{subsec:rl} and a simple-yet-effective rule-based grasping module in Section \ref{subsec:pick}. In particular, we utilize reinforcement learning to learn the former rearranging module for two main reasons. First is related to the cognitive limitations (e.g., lack of depth information in visualization) of the human operator when using a 2D screen to teleoperation. Furthermore, the contact-rich nature of the pushing task makes it hard to utilize the automated planning framework (e.g., TAMP \cite{Nam_20}).

%
% Learning how to perform non-prehensile pushing tasks
%
\subsection{Learning Non-prehensile Rearranging Tasks}
\label{subsec:rl}

% The objective of learning based pushing + % Why trajectory-based RL over PPO
Intuitively speaking, we want the learned policy to come up with the joint trajectories of a manipulator that can successfully rearrange the blocking objects to accomplish the picking task better. Furthermore, we do not want the policy to simply output a single trajectory but rather provide multiple (but plausible) trajectories as options to the teleoperator. To this end, we utilize a trajectory-based RL method named deep latent policy gradient (DLPG) \cite{Choi_19_DLPG} instead of instance-based RL such as proximal policy optimization (PPO) \cite{Schulman_17} or soft actor-critic (SAC) \cite{Haarnoja_18}. While with the presence of an accurate simulator and stochastic policy (e.g., a Gaussian policy), instance-based RL may provide multiple trajectories. However, this assumption does not hold anymore when executing in the real world, where the trajectory-based RL method can still be used in this scenario. Furthermore, DLPG has shown its strength in sample efficiency in locomotion domains \cite{Choi_19_DLPG}.

% State action 
The state-space consists of two pieces of informaton: the location of the target object in a $2$-dimensional $XY$ space and the depth information of the current scene. We utilize the transferred weights of the pre-trained convolutional autoencoder (CAE) as initial weights of convolutional neural network (CNN)\cite{lu_21} to map a depth image to a $10$-dimensional feature space. Specifically, we pretrain a convolutional autoencoder which maps a $38 \times 64$ depth image to a $10$-dimensional feature space with $10,000$ images sampled with random object placements. The output of the policy (i.e., action space) of DLPG defines a probability distribution over continuous end-effector trajectories using Gaussian random paths (GRPs) \cite{Choi_16} in a $2$-dimensional $XY$ space. GRPs parametrize the distribution with anchor points where we use $3$ anchor points. Once the end-effector trajectories are sampled, we solve inverse kinematics to compute the joint trajectories. 

% DLPG
In particular, DLPG defines the distribution over trajectories by $p(\mathbf{x}|\mathbf{z}, \mathbf{s})$ where $\mathbf{x}$ is a set of anchor points, $\mathbf{z}$ is a latent vector usually modeled by a Gaussian distribution, and $\mathbf{s}$ is a state vector consists of scene representation $\mathbf{c}$ and the target object position $\mathbf{g}$. In other words, by sampling $\mathbf{z}$ from $p(\mathbf{z})$, one can sample multiple trajectories even with a fixed scene which allows us to give multiple candidates to the user in the context of teleoperation. The overall network architecture is shown in Figure \ref{fig:network}.

%
% Figure: Reward
%
\begin{figure}[t!] % htp
 \centering
 \includegraphics[width=0.3\textwidth]{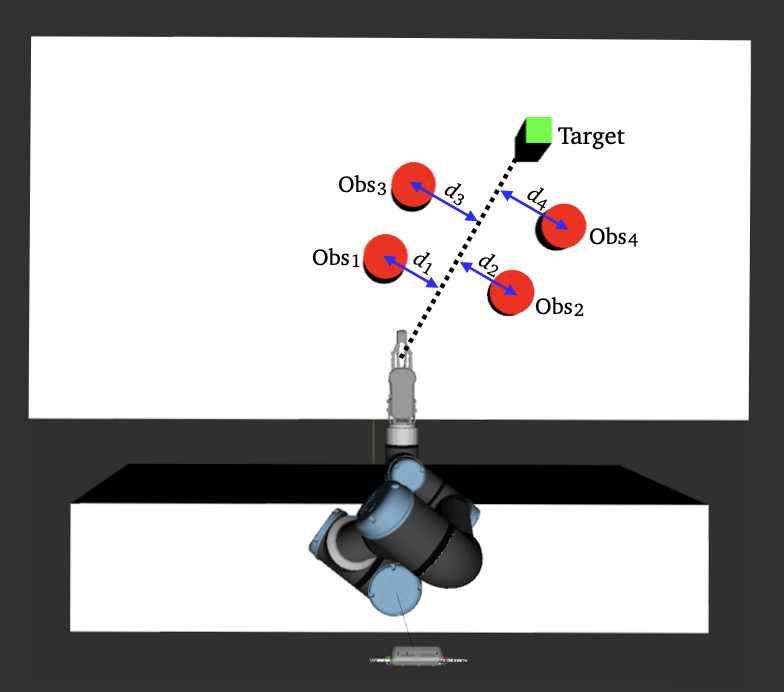}
 \vspace{-0.1pt}
 \caption{
 The Grasping margin reward for the rearranging task.
	}
	\label{fig:reward}
	\vspace{-0.1pt}
\end{figure}
%
% Simulation results
%
\begin{figure*}[t!]
 \centering
 \includegraphics[width=0.7\textwidth]{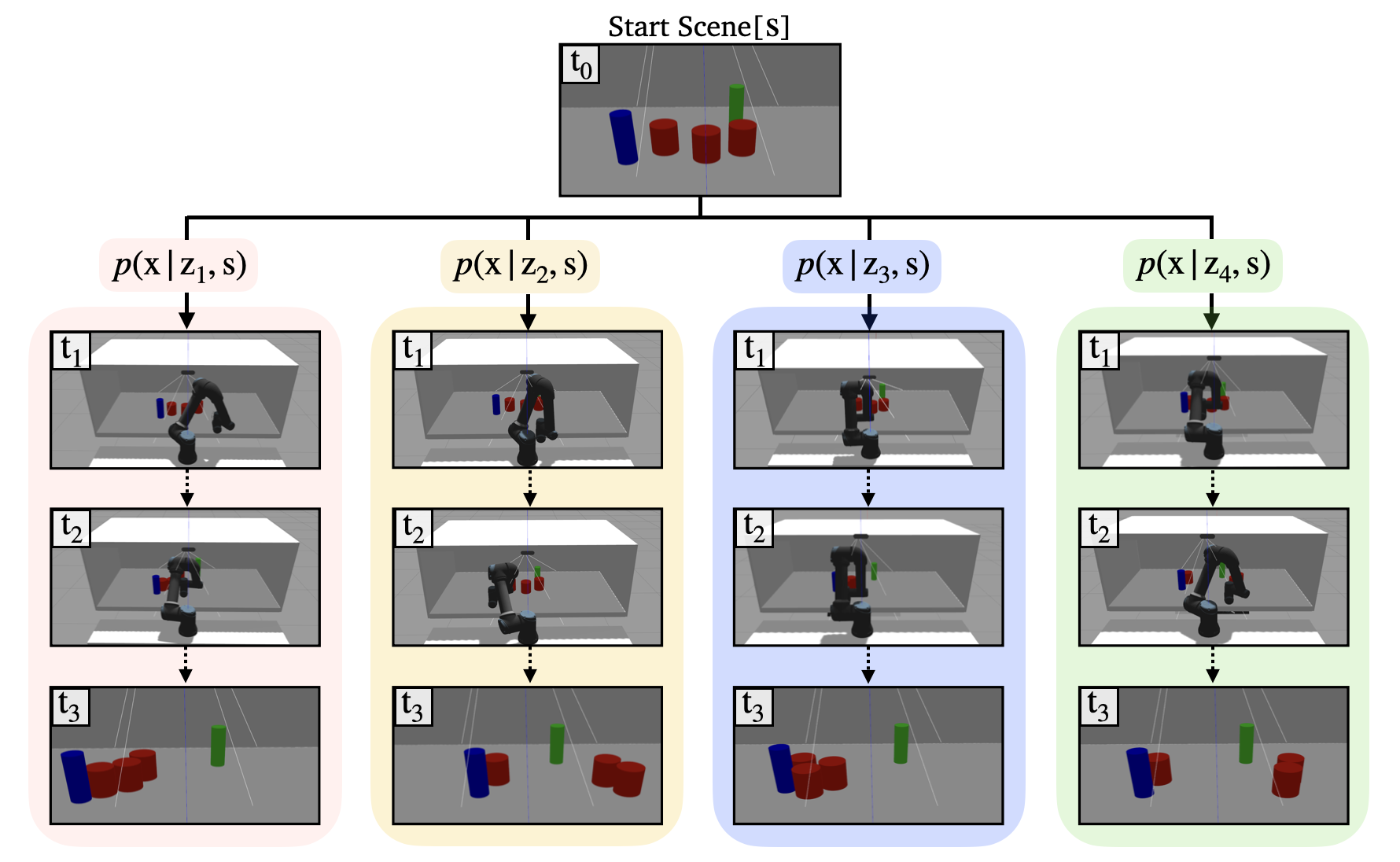}
 \vspace{-0.1pt}
 \caption{Different trajectories from the learned policy using trajectory-based reinforcement learning. Note that despite each trajectories are different and the resulting final scenes ($\mathrm{t_3}$) are more suitable for grasping than the start scene ($\mathrm{t_0}$). 
 }
 \label{fig:sampled_trajectories}
 \vspace{-0.1pt}
\end{figure*}
% Reward functions 
We carefully design two different reward functions. The first reward is related to how much the current non-prehensile rearrangement improves the direct grasping success rate. In other words, given $n$ obstacles in the scene, the line segment connecting the initial end-effector position of the manipulator and the target object. Then, we compute the minimum distances between the object and the line segment and compute the penalty of each object:
%
% Equation: margin reward
%
\begin{equation}
	r_{\text{margin}} =
	\begin{cases} 
	 100 d - 30, & \mbox{if }d \le 0.1 \text{m} \\
	 100 d - 25, & \mbox{if }d \le 0.2 \text{m} \\
	 100 d - 22.5, & \mbox{if }d \le 0.25 \text{m} \\
	 +15, & \text{otherwise} \\ 
	\end{cases}
	\label{eqn:r_margin}
\end{equation}
where $d$ is the minimum distance from the line segment to the center of the object. Once, $r_{\text{margin}}$ per each object is computed, we simply sum them to compute the total margin reward. Figure \ref{fig:reward} illustrates how this margin reward is computed. 

The second term is the safety reward that is to penalize the falling down of external objects. Since we are rearranging objects to better grasp the target object, it is prohibitive to aggressively sweep other objects. 
%
% Equation: safety reward
%
\begin{equation}
	r_{\text{safe}} =
	\begin{cases} 
		-30, & \mbox{if } \text{at least one object falls down} \\
		+10, & \mbox{if } \text{otherwise} 
	\end{cases}
	\label{eqn:r_safety}
\end{equation}
The total reward is computed by simply adding two reward functions in (\ref{eqn:r_margin}) and (\ref{eqn:r_safety}).

% Domain randomize
For the learned policy to generalize better to an unseen environment, we randomly place a random number of objects in the workspace \cite{Tobin_17}. Specifically, we place five to seven objects including a target object in the scene at random locations within the table. The objects include thin ones that are more likely to fall when touched by a manipulator. Since we utilize the depth image as an input and penalize the reward when any object falls down, the learned motion for the non-prehensile mode will likely to rearrange objects that do not fall.

%
% Picking object
%
\subsection{Picking up the object}
\label{subsec:pick}

We implement a simple yet effective grasping algorithm from scratch. We utilize a density-based spatial clustering of applications with noise (DBSCAN) \cite{Birant_07} for clustering objects using point cloud data obtained from an RGB-D camera. Once objects are clustered, we publish them on a ROS visualization tool named RViz. Then, the teleoperator can choose one of the objects as the target object among clusters by using the point stamp function of RViz. The main advantage of the approach is that even when the user does not precisely select the center of the target object, the center of the object is selected as the target position using the clustered data. 

% How to implement pick up
Once the target object to grasp is selected by the user, the algorithm first draws a line segment from the initial position of the end-effector to the center of the object. From this line segment, the grasp pose is estimated, and the intermediate end-effector positions are also computed by linear interpolation.

%
% Depth image
% 
\begin{figure*}[t!] 
	\centering
	\subfigure[]{
	\includegraphics[width=0.3\linewidth]{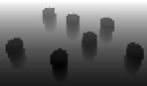}
	}
	\subfigure[]{
	\includegraphics[width=0.3\linewidth]{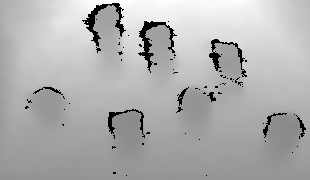}
	}
	\subfigure[]{
	\includegraphics[width=0.3\linewidth]{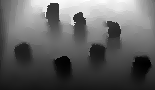}
	}
	\caption{Different depth images obtained from (a) the simulational environment, (b) real-world environment, and (c) the post-processed depth image.}
	\label{fig:depth}
	\vspace{-0.1pt}
\end{figure*}
%
% Experiment
%
\section{Experiment} \label{sec:experiment}
%
% Experimental Detail
%
\subsection{Implementation Detail} \label{subsec:exp_detail}

% Basic setup
In both simulations and real-world experiments, a 6-DoF UR5e manipulator with an OnRobot RG2 gripper equipped with an Intel RealSense D435 sensor is utilized. For control and simulation purposes, we use Robot Operating System (ROS) with a Gazebo simulator mainly due to simulating the RGB-D sensor. For the sake of efficient rollout during the reinforcement learning, Ray \cite{Moritz_18}, a well-known package for a distributed system, is further utilized. 

% Speed
The publishing rate of the joint positions is set to $500$HZ, a default setting for a UR5e manipulator. Once the end-effector trajectory is computed from either the learning-based non-prehensile manipulation module in Section \ref{subsec:rl} or grasping module in Section \ref{subsec:pick}, the timestamps are recomputed and interpolated so that the velocity of the end-effector is fixed to $0.1$m/s. We observe that this simple time-rescaling increases the safety and stability of the control in real-world experiments.  

%
% Reward graph
%
\begin{figure}[t!]
 \centering
 \includegraphics[width=0.4\textwidth]{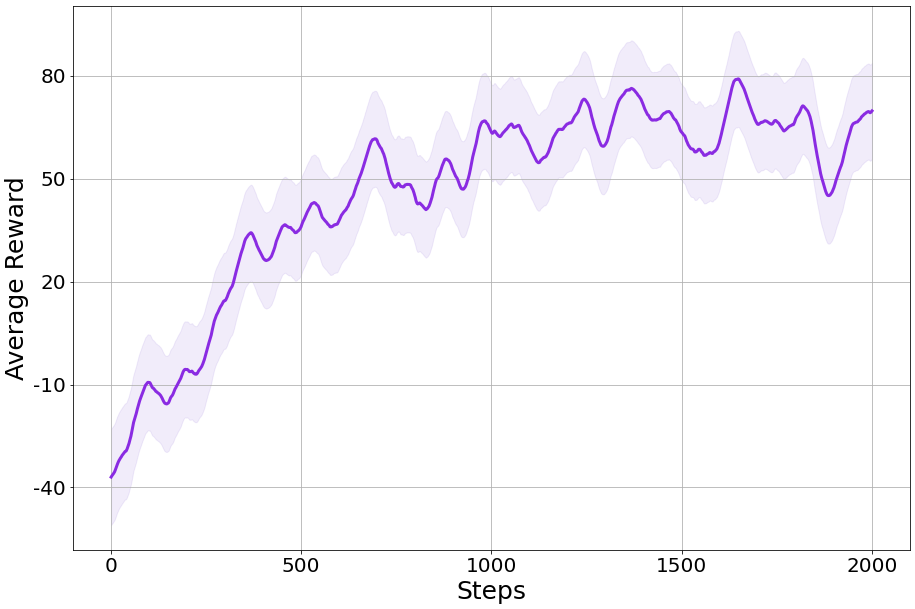}
 \vspace{-0.1pt}
 \caption{
 Average reward computed with a moving average ($50$-steps) where $40$ distributed agents are utilized using Ray \cite{Moritz_18}.} 
	\label{fig:reward_result}
	\vspace{-0.1pt}
\end{figure}
%
% User study Result
% 
\begin{figure}[t] %%% 
	\centering
	\subfigure[]{
	\includegraphics[width=0.85\linewidth]{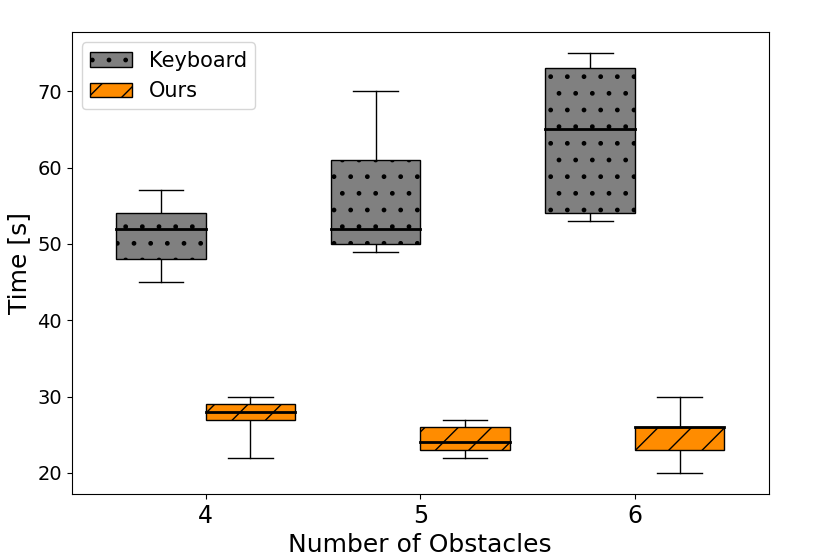}
	\label{fig:userstudy_time}
	}
	\subfigure[]{
	\includegraphics[width=0.85\linewidth]{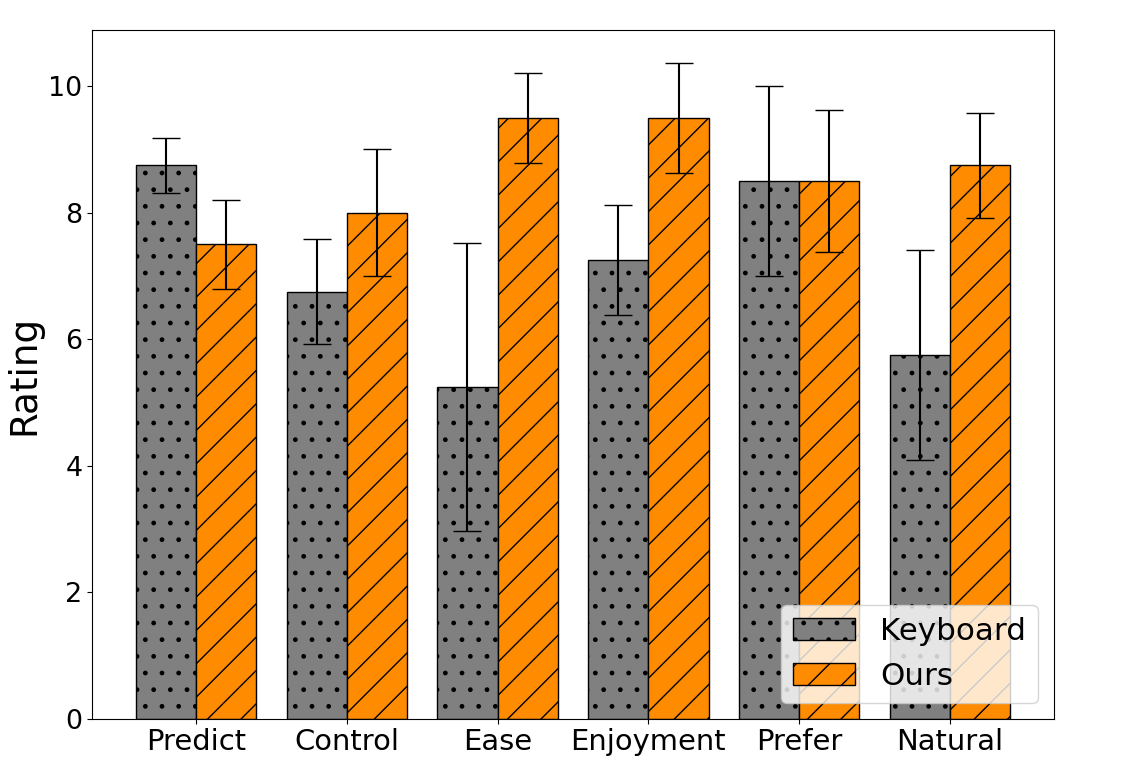}
	\label{fig:userstudy_preference}
	}
	\caption{User study results of the real-world experiments.}
	\label{fig:userstudy}
\end{figure}
%
% Simulational Results
%
\subsection{Simulational Results} \label{subsec:exp_sim}

Figure \ref{fig:reward_result} shows the rewards of the proposed method during the training phase, where we can see that the proposed method successfully trained the policy for the non-prehensile manipulation task. We would like to emphasize that the policy should be able to provide multiple trajectory candidates for the user to select. Figure \ref{fig:sampled_trajectories} shows four different rearranging trajectories of a manipulator given the same scene by changing the latent vector in DLPG. While the different latent vectors provide different behaviors, the results of the non-prehensile manipulation are all effective in terms of performing grasping. 

%
% Real-world Result
%
\subsection{Real-world Experiments and User Study} \label{subsec:exp_real}

\begin{figure*}[hbt!] %%% 
\centering
\subfigure[]{
\includegraphics[width=0.26\linewidth]{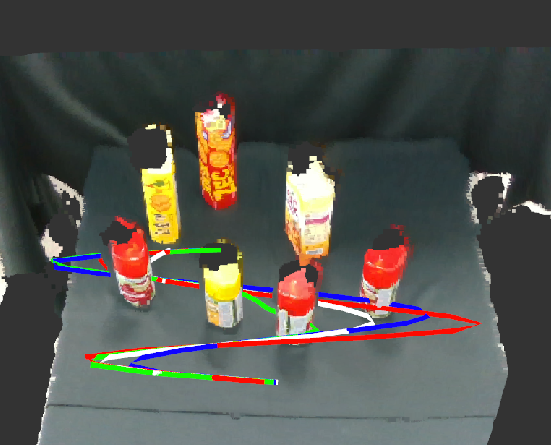}
}
\subfigure[]{
\includegraphics[width=0.256\linewidth]{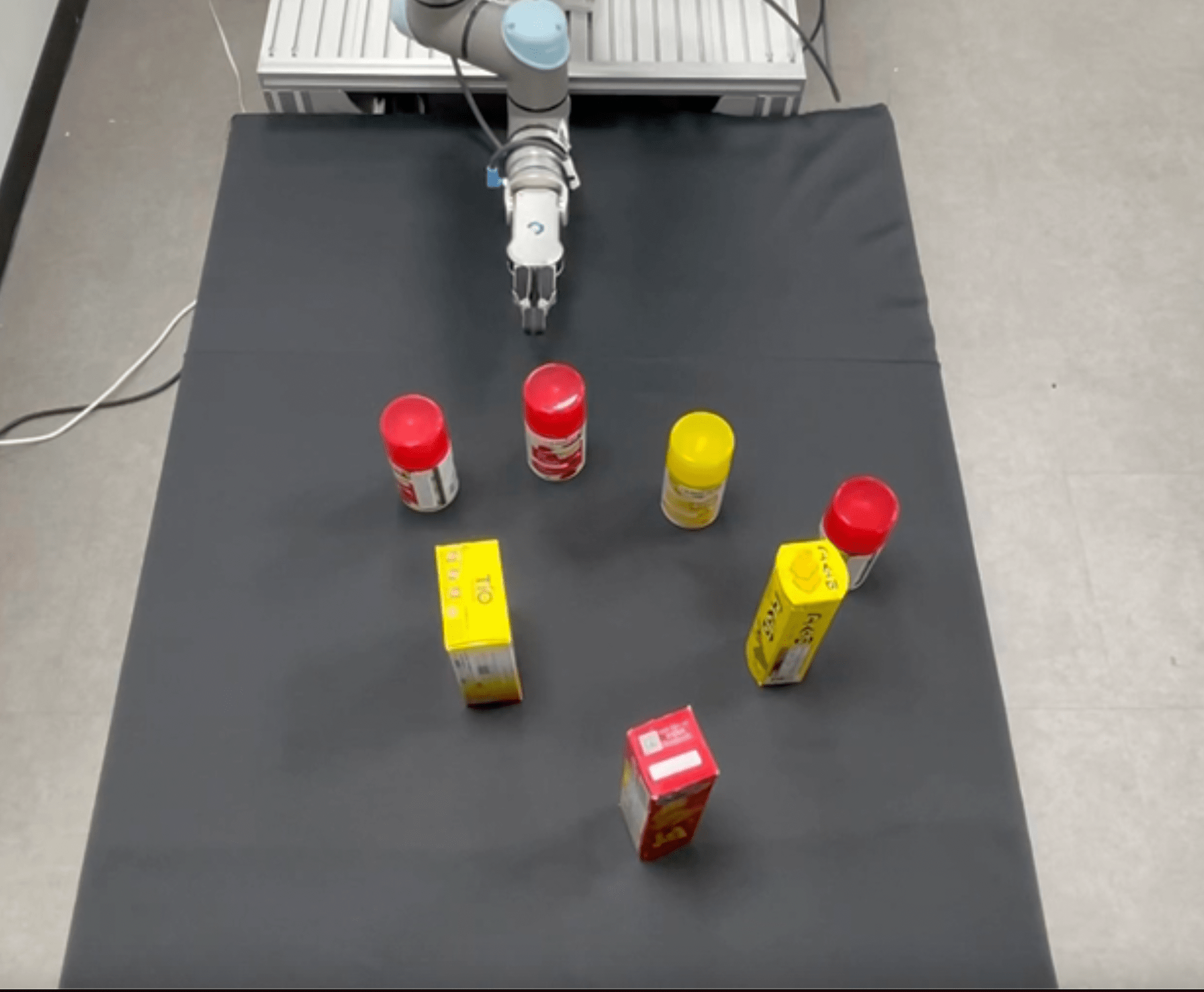}
}
%  \subfigure[]{
%  \includegraphics[width=0.26\linewidth]{figs/push2_min.png}
%  }
\subfigure[]{
\includegraphics[width=0.256\linewidth]{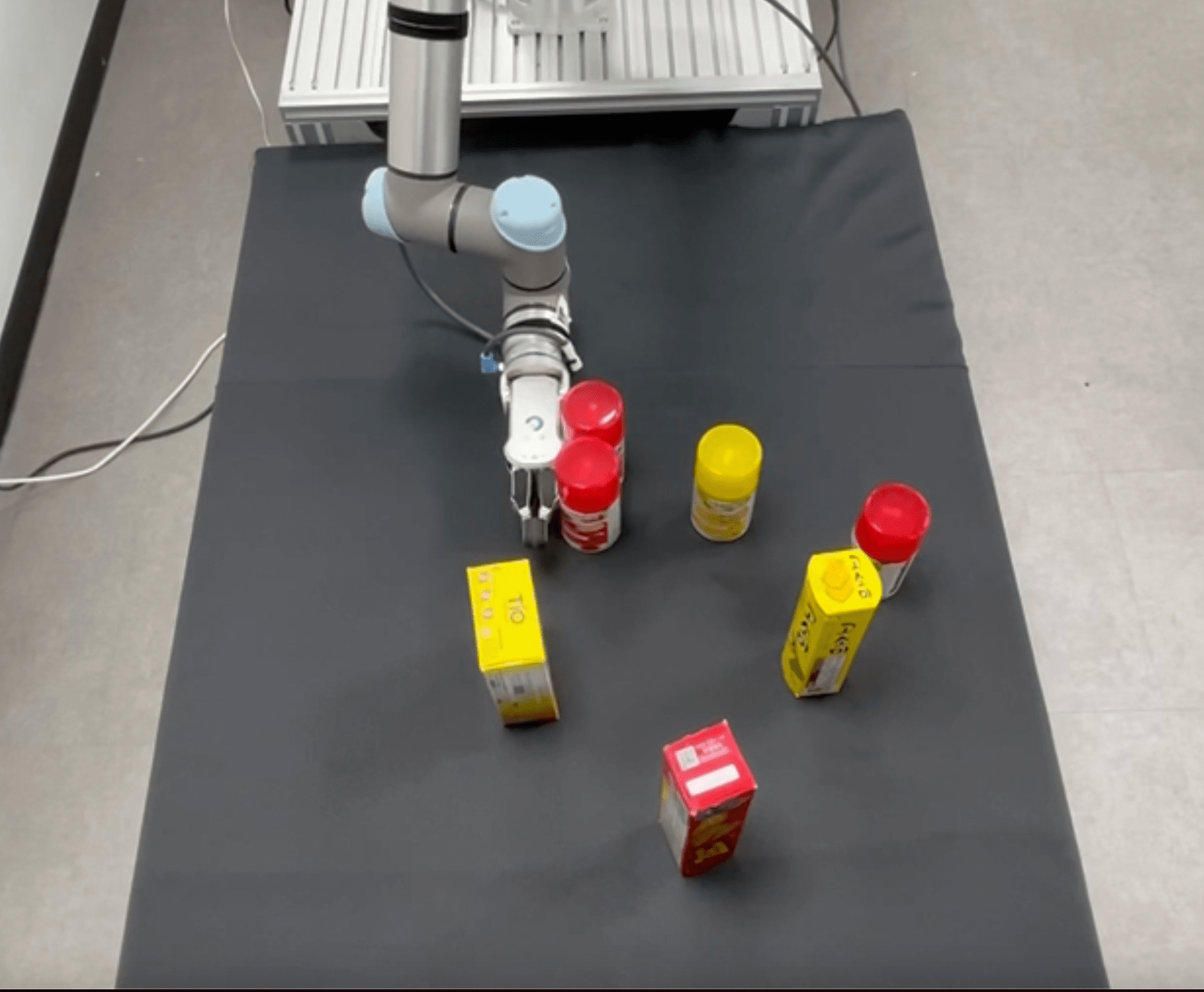}
}
\subfigure[]{
\includegraphics[width=0.256\linewidth]{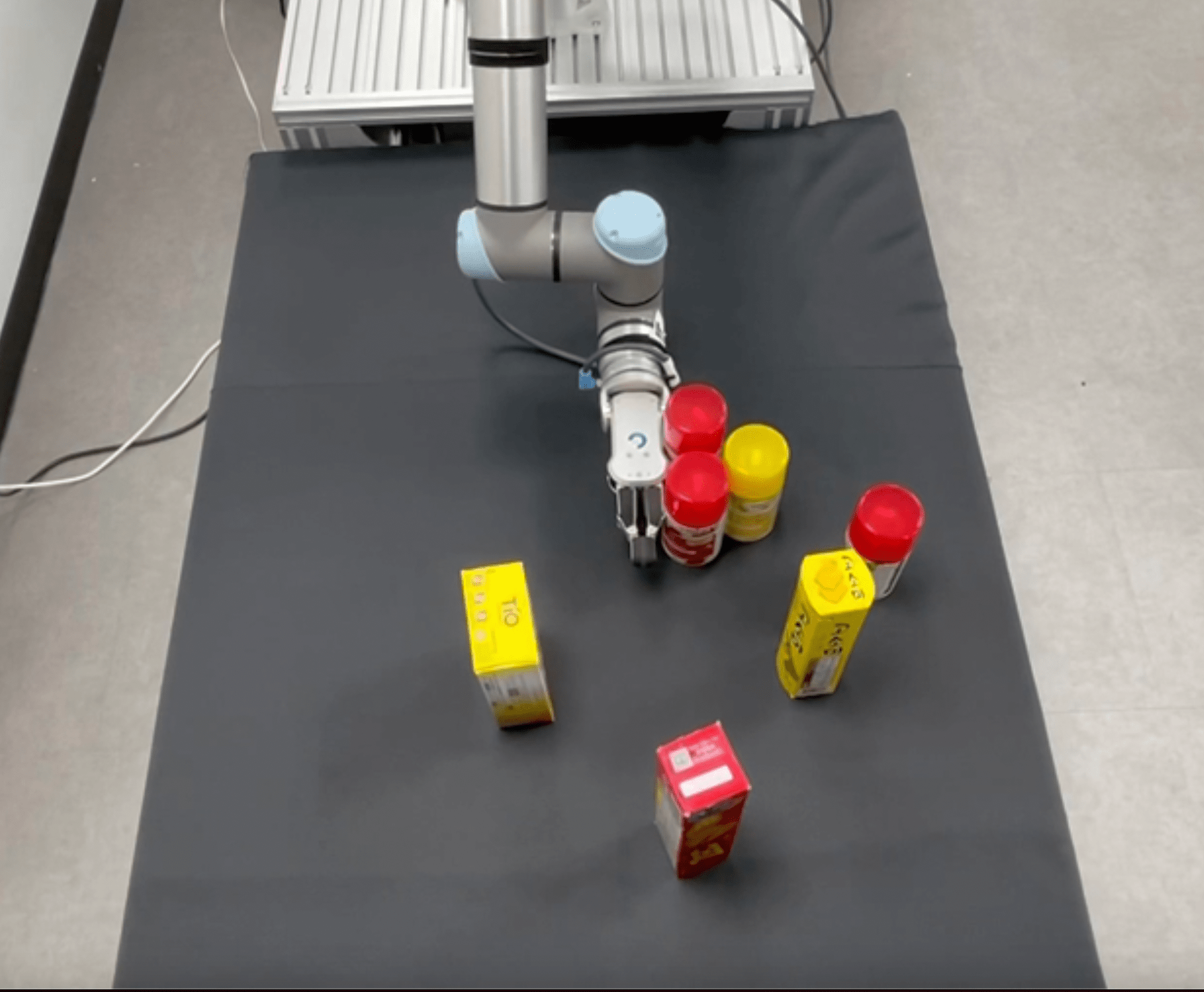}
}
\subfigure[]{
\includegraphics[width=0.256\linewidth]{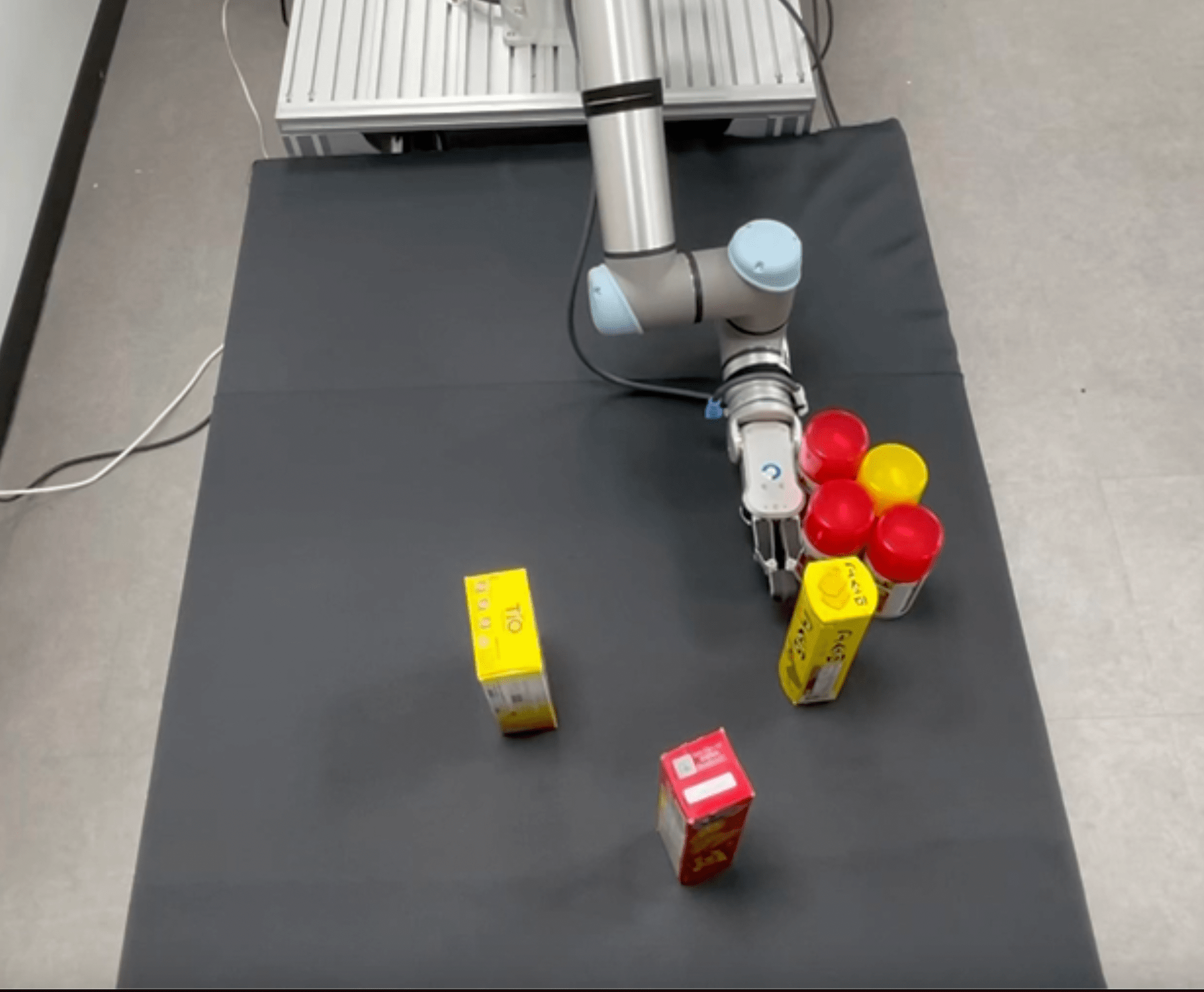}
}
\subfigure[]{
\includegraphics[width=0.256\linewidth]{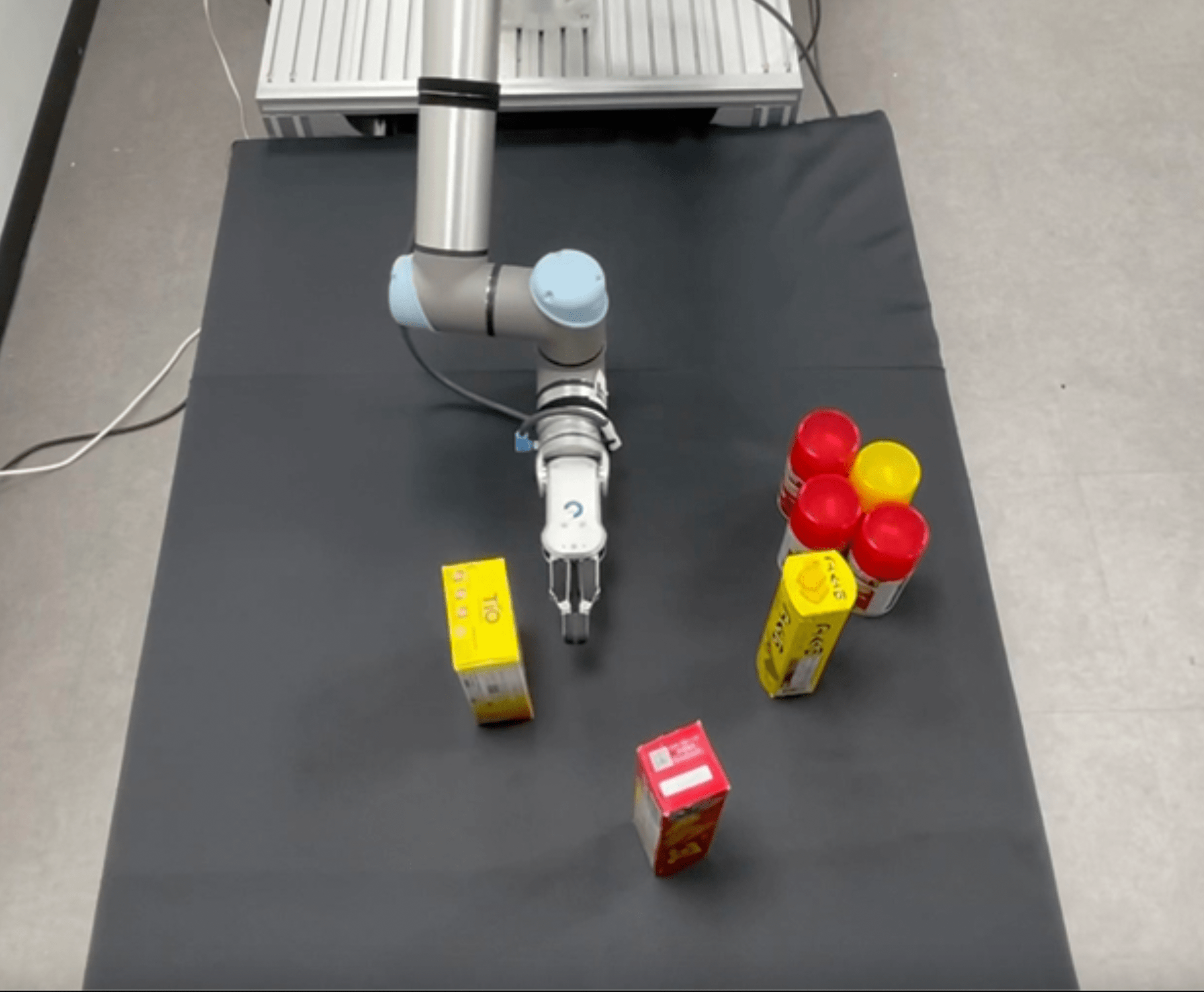}
}
\subfigure[]{
\includegraphics[width=0.256\linewidth]{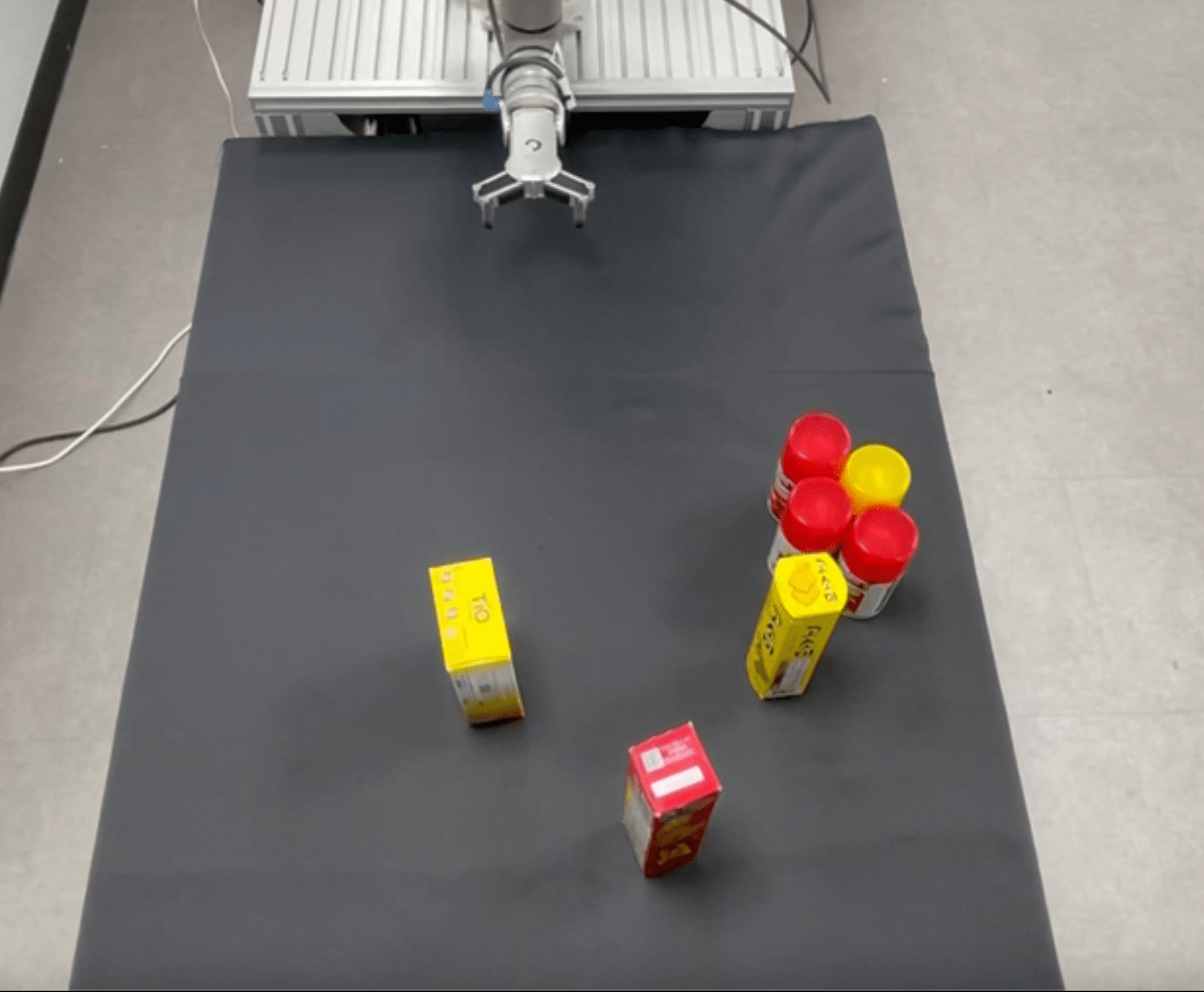}
}
\subfigure[]{
\includegraphics[width=0.256\linewidth]{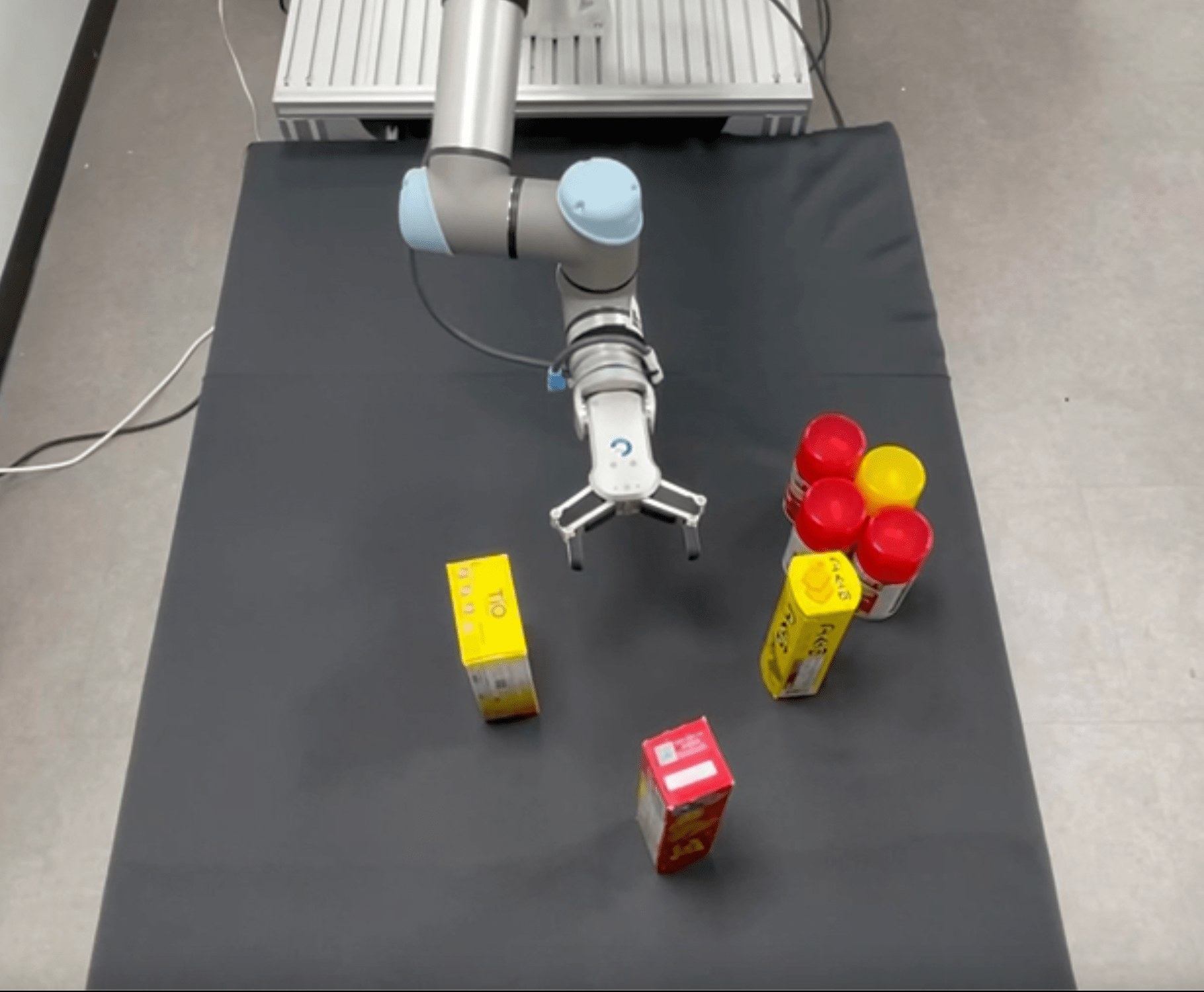}
}
\subfigure[]{
\includegraphics[width=0.256\linewidth]{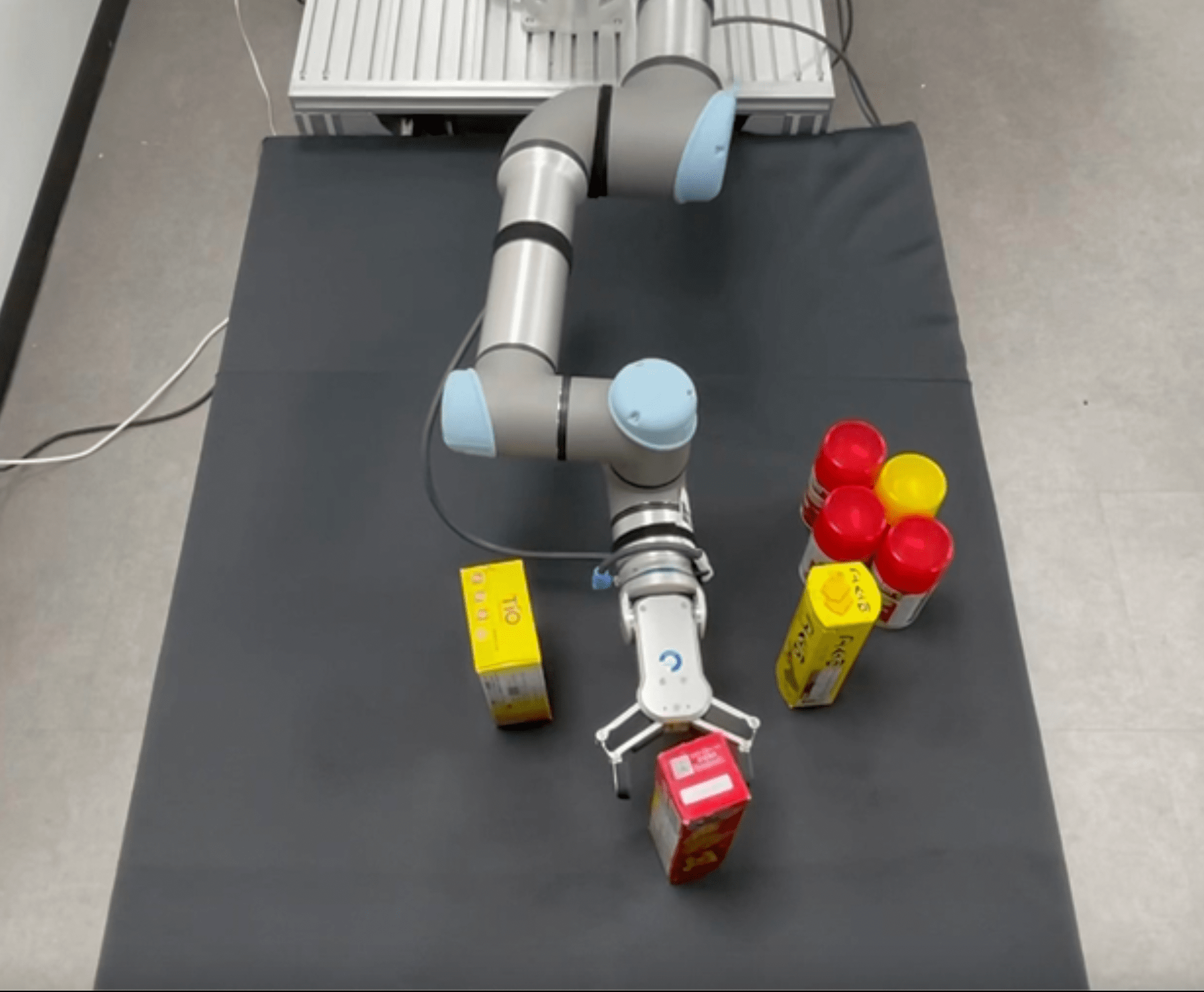}
}
\subfigure[]{
\includegraphics[width=0.26\linewidth]{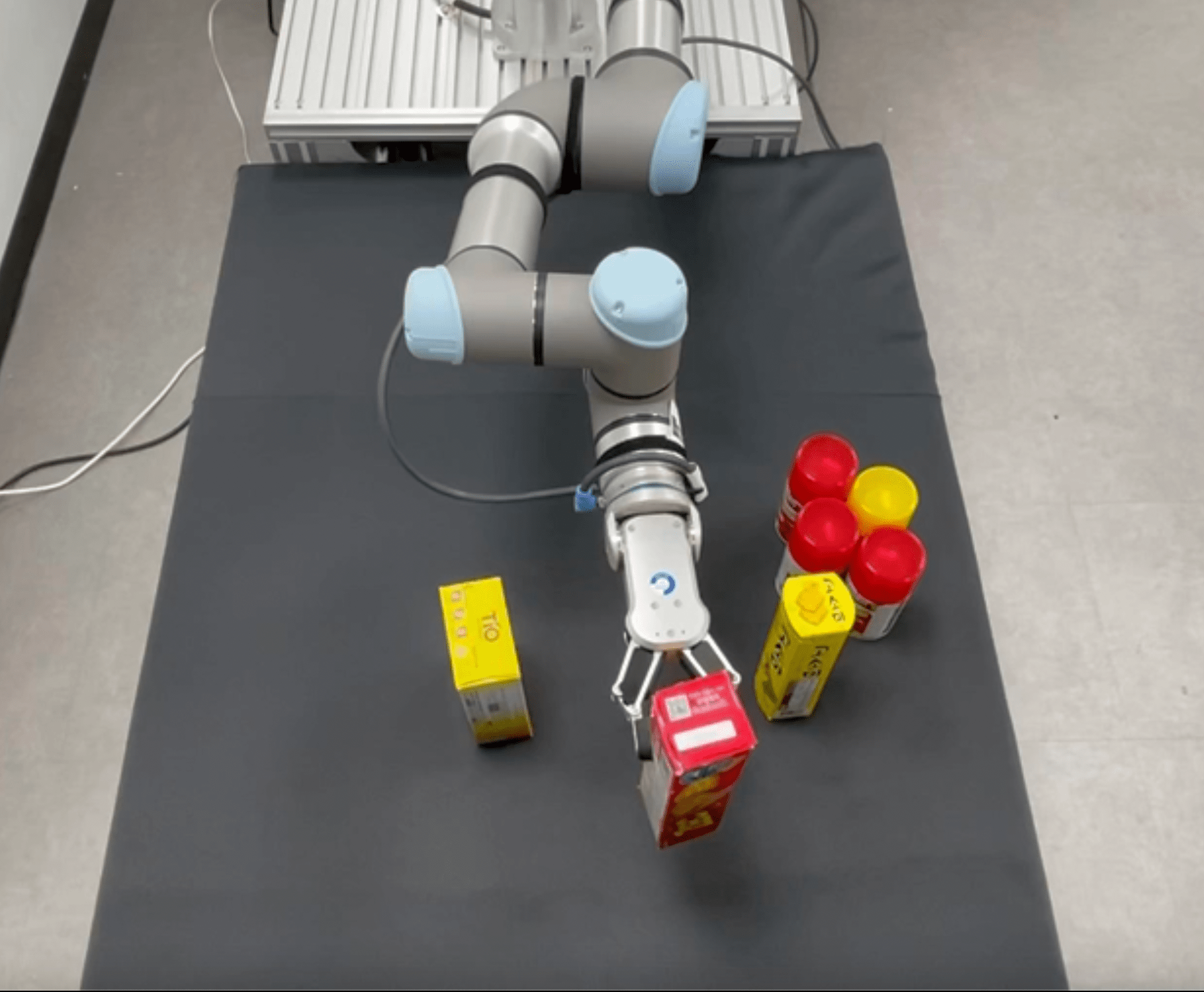}
}
\subfigure[]{
\includegraphics[width=0.26\linewidth]{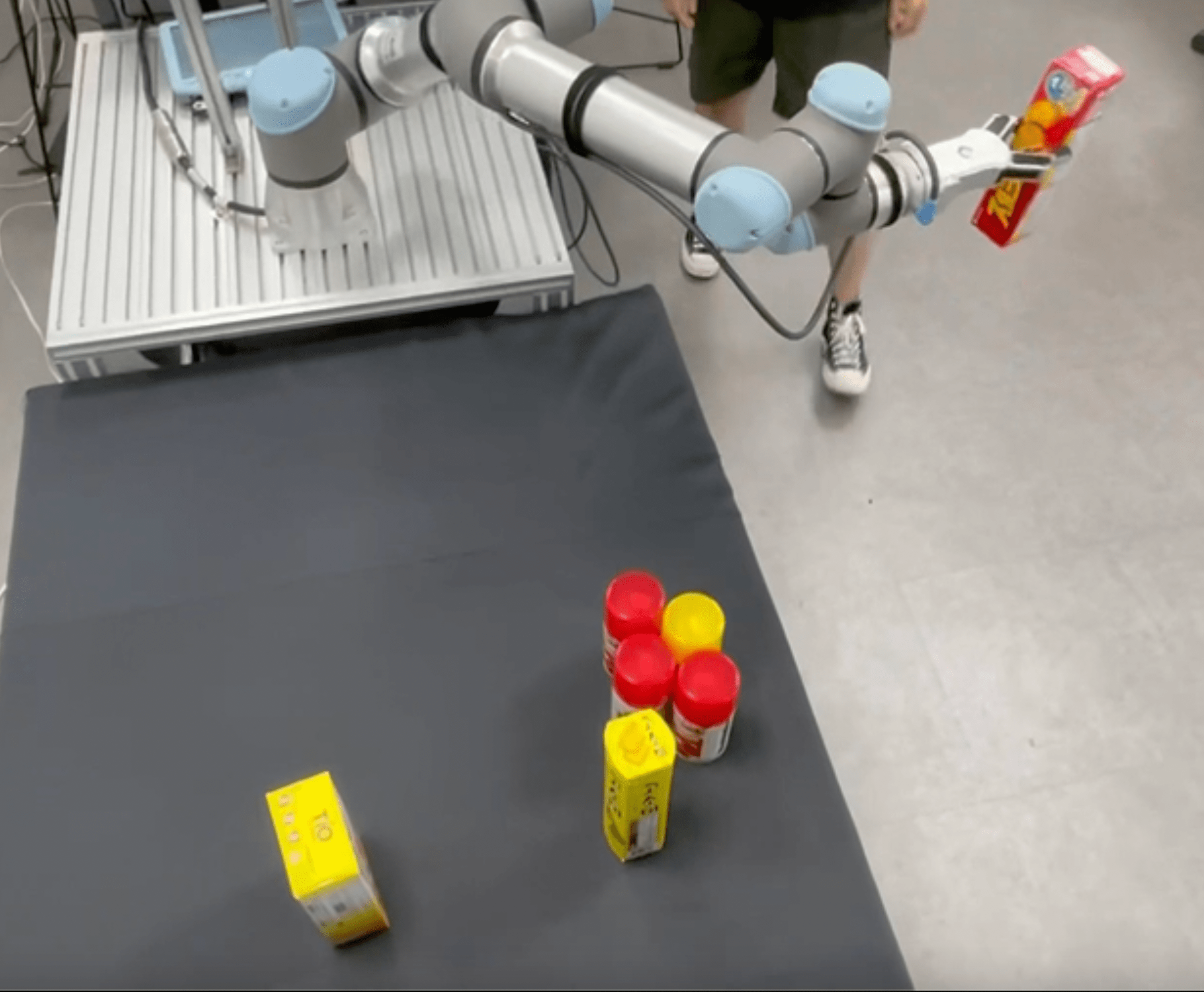}
}
\subfigure[]{
\includegraphics[width=0.256\linewidth]{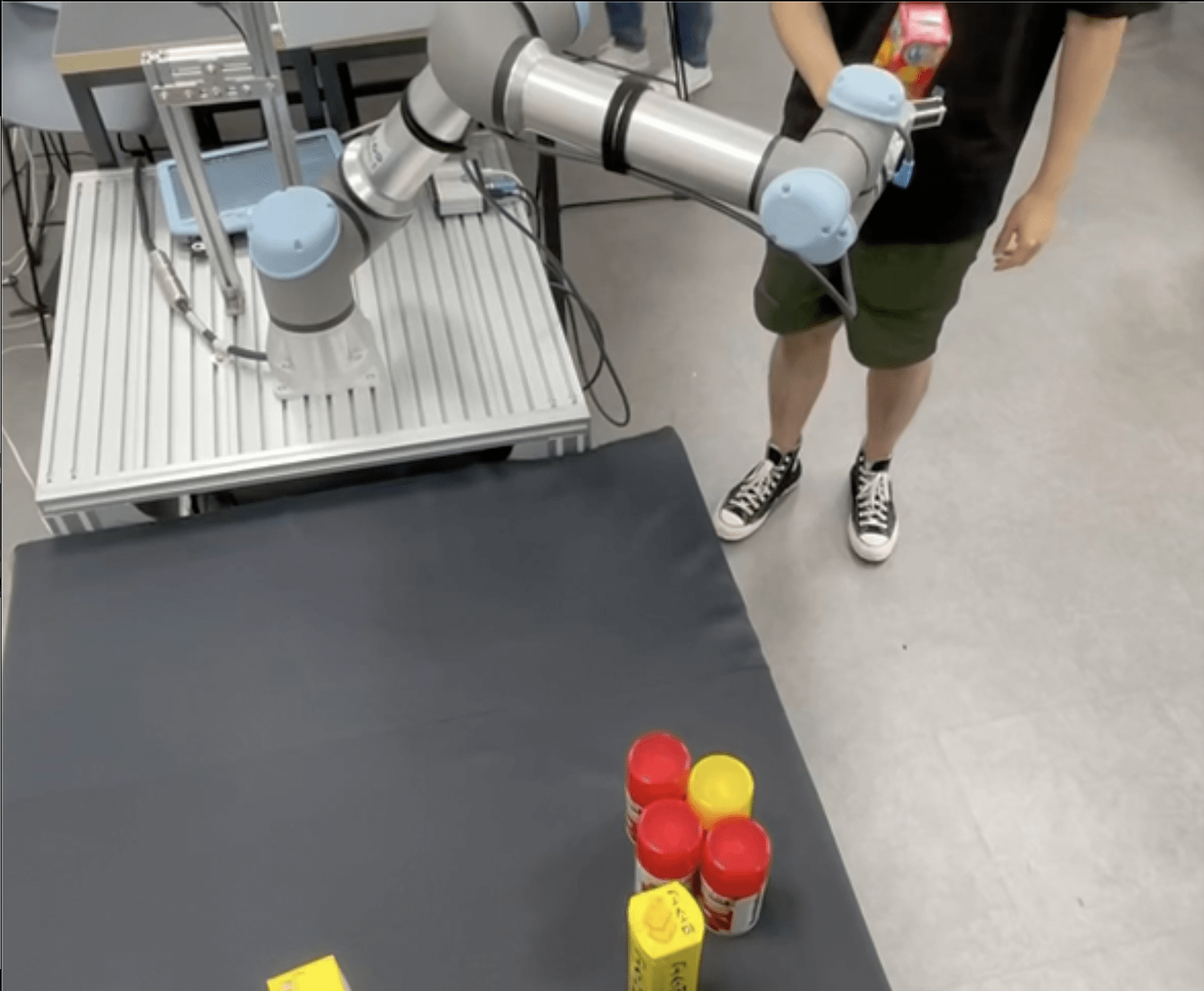}
}
\caption{
Snapshots of (a-f) the non-prehensile manipulation mode where the red box is selected as the target object to grasp and (g-l) the grasping mode. Note that the generated joint trajectories from the learned policy successfully rearranged other objects so that direct grasping becomes possible. The RViz interface for selecting the sweeping trajectory is shown in (a), where different sampled trajectories are shown with different colors. 
}
\label{fig:snapshot}
\end{figure*}

% Sim-to-real
We use the trained policy in Section \ref{subsec:rl} directly in the real-world experiments without retraining (i.e., sim2real). However, there exists some level of discrepancies between the perception performances of the simulation environment (i.e., Gazebo) and real-world sensing. To mitigate this gap, we utilize the depth image post-processing method for Intel RealSense SDK using PyRealSense. Figure \ref{fig:depth} shows the depth images obtained from the simulation environment, real-world, and post-processing, respectively. The snapshots of both sweeping motion and pick-and-place motion are shown in Figure \ref{fig:snapshot}. 

% User evaluation
We conduct user studies to evaluate the effectiveness of the proposed semi-autonomous teleoperation framework in terms of time efficiencies and user preferences. For the user study, we recruited five volunteers that provided informed consent (age range  27.8 ± 3.5), with all participants having no prior robot teleoperation experience. We assume that the target object is blocked by a number of other objects varying from $4$ to $6$. In particular, the proposed method is compared with a manual control proposed in \cite{Freire_00} using a keyboard where the user can move the end-effector into four different directions (i.e., forward, backward, right, and left).

% Results
Figure \ref{fig:userstudy_time} shows the time duration of the non-prehensile rearranging mode without prehensile grasping mode while varying the number of blocking objects. In all different cases, our proposed method outperforms the baseline manual control in terms of time efficiency. We would like to emphasize that while the time duration using the baseline method increases as the number of objects increases, our proposed method is less affected by the number of obstacles in that we leverage sweep-like motions generated from the learned policy. 

% User preference
Furthermore, we conduct other user studies regarding six different factors in terms of the predictability, controllability, easiness, enjoyment, preferences, and naturalness of the teleoperation process. Note that these factors are selected from \cite{Losey_20}. Figure \ref{fig:userstudy_preference} illustrates the comparative results of the size different factors. First, the manual control shows its strength in terms of predictability as the users can directly control the end-effector to their desired direction. However, our proposed method shows superior performances in the remaining factors. Specifically, it outperforms the baseline in terms of easiness which is related to the time duration for picking the target objects. It also shows less standard deviation compared to the baseline, which indicates manual control is more likely to depend on personal skill. % So, we believe that our approach will make more users enable controlling a robot with less burden.

%
% Snapshots of pushing and grasping
%
%
% Conclusion
%
\section{Conclusion} \label{sec:conclusion}

In this paper, we have proposed a semi-autonomous teleoperation framework that effectively combines a learning-based non-prehensile manipulation task and a prehensile grasping task. In particular, we assume that the target object is located in a cluttered environment where a depth image is observed during the teleoperation. To effectively generated the sweeping motion, we utilize trajectory-based reinforcement learning where the learned policy can generate multiple trajectories by sampling the latent vector. Then, the user selects a preferred trajectory. The proposed method has been evaluated in both simulations and real-world environments, where it has shown its strength in time efficiency. The user study tells us that the users can not only pick up the target object in a shorter duration but also are more satisfied with our framework in terms of controllability, easiness, and naturalness of the motion. We believe that this is due to our semi-autonomous framework that can provide multiple high-level options to the user.
\balance

%
% 						Reference 
%
\bibliographystyle{IEEEtran}
\bibliography{references}

\begin{thebibliography}{10}
\providecommand{\url}[1]{#1}
\csname url@samestyle\endcsname
\providecommand{\newblock}{\relax}
\providecommand{\bibinfo}[2]{#2}
\providecommand{\BIBentrySTDinterwordspacing}{\spaceskip=0pt\relax}
\providecommand{\BIBentryALTinterwordstretchfactor}{4}
\providecommand{\BIBentryALTinterwordspacing}{\spaceskip=\fontdimen2\font plus
\BIBentryALTinterwordstretchfactor\fontdimen3\font minus
  \fontdimen4\font\relax}
\providecommand{\BIBforeignlanguage}[2]{{%
\expandafter\ifx\csname l@#1\endcsname\relax
\typeout{** WARNING: IEEEtran.bst: No hyphenation pattern has been}%
\typeout{** loaded for the language `#1'. Using the pattern for}%
\typeout{** the default language instead.}%
\else
\language=\csname l@#1\endcsname
\fi
#2}}
\providecommand{\BIBdecl}{\relax}
\BIBdecl

\bibitem{Schilling_19}
M.~Schilling, W.~Burgard, K.~Muelling, B.~Wrede, and H.~Ritter, ``Shared
  autonomy---learning of joint action and human-robot collaboration,''
  \emph{Frontiers in neurorobotics}, vol.~13, p.~16, 2019.

\bibitem{Losey_20}
D.~P. Losey, K.~Srinivasan, A.~Mandlekar, A.~Garg, and D.~Sadigh, ``Controlling
  assistive robots with learned latent actions,'' in \emph{Proc. of IEEE
  International Conference on Robotics and Automation (ICRA)}.\hskip 1em plus
  0.5em minus 0.4em\relax IEEE, 2020, pp. 378--384.

\bibitem{Jeon_20}
H.~J. Jeon, D.~P. Losey, and D.~Sadigh, ``Shared autonomy with learned latent
  actions,'' \emph{arXiv preprint arXiv:2005.03210}, 2020.

\bibitem{Karamcheti_21}
S.~Karamcheti, A.~J. Zhai, D.~P. Losey, and D.~Sadigh, ``Learning visually
  guided latent actions for assistive teleoperation,'' in \emph{Learning for
  Dynamics and Control}.\hskip 1em plus 0.5em minus 0.4em\relax PMLR, 2021, pp.
  1230--1241.

\bibitem{Lee_19}
J.~Lee, Y.~Cho, C.~Nam, J.~Park, and C.~Kim, ``Efficient obstacle rearrangement
  for object manipulation tasks in cluttered environments,'' in \emph{Proc. of
  IEEE International Conference on Robotics and Automation (ICRA)}.\hskip 1em
  plus 0.5em minus 0.4em\relax IEEE, 2019, pp. 183--189.

\bibitem{Lee_21}
J.~Lee, C.~Nam, J.~H. Park, and C.~Kim, ``Tree search-based task and motion
  planning with prehensile and non-prehensile manipulation for obstacle
  rearrangement in clutter,'' in \emph{Proc. of IEEE International Conference
  on Robotics and Automation (ICRA)}.\hskip 1em plus 0.5em minus 0.4em\relax
  IEEE, 2021, pp. 8516--8522.

\bibitem{papallas_20}
R.~Papallas and M.~R. Dogar, ``Non-prehensile manipulation in clutter with
  human-in-the-loop,'' in \emph{Proc. of the IEEE International Conference on
  Robotics and Automation (ICRA)}.\hskip 1em plus 0.5em minus 0.4em\relax IEEE,
  2020, pp. 6723--6729.

\bibitem{king_17}
J.~E. King, V.~Ranganeni, and S.~S. Srinivasa, ``Unobservable monte carlo
  planning for nonprehensile rearrangement tasks,'' in \emph{Proc. of the IEEE
  International Conference on Robotics and Automation (ICRA)}.\hskip 1em plus
  0.5em minus 0.4em\relax IEEE, 2017, pp. 4681--4688.

\bibitem{yuan_18}
W.~Yuan, J.~A. Stork, D.~Kragic, M.~Y. Wang, and K.~Hang, ``Rearrangement with
  nonprehensile manipulation using deep reinforcement learning,'' in
  \emph{Proc. of the IEEE International Conference on Robotics and Automation
  (ICRA)}.\hskip 1em plus 0.5em minus 0.4em\relax IEEE, 2018, pp. 270--277.

\bibitem{Schulman_15}
J.~Schulman, S.~Levine, P.~Abbeel, M.~Jordan, and P.~Moritz, ``Trust region
  policy optimization,'' in \emph{Proc. of IEEE International Conference on
  Machine Learning (ICML)}.\hskip 1em plus 0.5em minus 0.4em\relax PMLR, 2015,
  pp. 1889--1897.

\bibitem{Schulman_17}
J.~Schulman, F.~Wolski, P.~Dhariwal, A.~Radford, and O.~Klimov, ``Proximal
  policy optimization algorithms,'' \emph{arXiv preprint arXiv:1707.06347},
  2017.

\bibitem{Haarnoja_18}
T.~Haarnoja, A.~Zhou, P.~Abbeel, and S.~Levine, ``Soft actor-critic: Off-policy
  maximum entropy deep reinforcement learning with a stochastic actor,'' in
  \emph{Proc. of IEEE International Conference on Machine Learning
  (ICML)}.\hskip 1em plus 0.5em minus 0.4em\relax PMLR, 2018, pp. 1861--1870.

\bibitem{Choi_19_DLPG}
S.~Choi and J.~Kim, ``Trajectory-based probabilistic policy gradient for
  learning locomotion behaviors,'' in \emph{Proc. of IEEE International
  Conference on Robotics and Automation (ICRA)}.\hskip 1em plus 0.5em minus
  0.4em\relax IEEE, 2019, pp. 1--7.

\bibitem{Freire_00}
E.~Freire, R.~Vassallo, R.~Alves, and T.~Bastos-Filho, ``Teleoperation of a
  mobile robot through the internet,'' in \emph{Proc. of the 43rd Midwest
  Symposium on Circuits and Systems (MWSCAS)}, 2000.

\bibitem{Nam_20}
C.~Nam, J.~Lee, S.~H. Cheong, B.~Y. Cho, and C.~Kim, ``Fast and resilient
  manipulation planning for target retrieval in clutter,'' in \emph{Proc. of
  IEEE International Conference on Robotics and Automation (ICRA)}.\hskip 1em
  plus 0.5em minus 0.4em\relax IEEE, 2020, pp. 3777--3783.

\bibitem{Liu_13}
\BIBentryALTinterwordspacing
Y.-C. Liu and N.~Chopra, ``Control of semi-autonomous teleoperation system with
  time delays,'' \emph{Automatica}, vol.~49, no.~6, pp. 1553--1565, 2013.
  [Online]. Available:
  \url{https://www.sciencedirect.com/science/article/pii/S0005109813000733}
\BIBentrySTDinterwordspacing

\bibitem{Choi_16}
S.~Choi, K.~Lee, and S.~Oh, ``Gaussian random paths for real-time motion
  planning,'' in \emph{Proc. of IEEE/RSJ International Conference on
  Intelligent Robots and Systems (IROS)}.\hskip 1em plus 0.5em minus
  0.4em\relax IEEE, 2016, pp. 1456--1461.

\bibitem{Tobin_17}
J.~Tobin, R.~Fong, A.~Ray, J.~Schneider, W.~Zaremba, and P.~Abbeel, ``Domain
  randomization for transferring deep neural networks from simulation to the
  real world,'' in \emph{Proc. of IEEE/RSJ international conference on
  Intelligent Robots and Systems (IROS)}.\hskip 1em plus 0.5em minus
  0.4em\relax IEEE, 2017, pp. 23--30.

\bibitem{Birant_07}
D.~Birant and A.~Kut, ``St-dbscan: An algorithm for clustering
  spatial--temporal data,'' \emph{Data \& knowledge engineering}, vol.~60,
  no.~1, pp. 208--221, 2007.

\bibitem{Moritz_18}
P.~Moritz, R.~Nishihara, S.~Wang, A.~Tumanov, R.~Liaw, E.~Liang, M.~Elibol,
  Z.~Yang, W.~Paul, M.~I. Jordan \emph{et~al.}, ``Ray: A distributed framework
  for emerging ai applications,'' in \emph{Symposium on Operating Systems
  Design and Implementation (OSDI)}, 2018, pp. 561--577.

\end{thebibliography}

\end{document}